\documentclass[lettersize,journal]{IEEEtran}
\usepackage{amsmath,amsfonts}
\usepackage{algorithm}
\usepackage{array}
\usepackage[caption=false,font=normalsize,labelfont=sf,textfont=sf]{subfig}
\usepackage{textcomp}
\usepackage{stfloats}
\usepackage{url}
\usepackage{verbatim}
\usepackage{graphicx}
\usepackage{cite}
\usepackage{booktabs}   
\usepackage{multirow}   
\usepackage{makecell}   
\usepackage{bm} 
\usepackage{setspace}
\usepackage{array}      
\usepackage{pifont} 
\usepackage{algpseudocode}
\usepackage{amsmath, amssymb, bm}

\begin{document}

\title{\huge Diffusion$^\mathbf{2}$: Dual Diffusion Model with Uncertainty-Aware Adaptive Noise for Momentary Trajectory Prediction}

\author{Yuhao~Luo$^\dagger$,~\IEEEmembership{Graduate Student Member,~IEEE},      
Yuang~Zhang$^\dagger$,
Kehua~Chen,~\IEEEmembership{Member,~IEEE},
Xinyu~Zheng, Shucheng~Zhang, Sikai~Chen and
Yinhai~Wang, ~\IEEEmembership{Fellow,~IEEE}

\thanks{$^\dagger$ Yuhao Luo and Yuang Zhang contributed equally to this work.}
\thanks{Yuhao Luo and Sikai Chen are with Civil and Environmental Engineering Department, University of Wisconsin-Madison, Madison, USA}
\thanks{Yuang Zhang, Kehua Chen, Shucheng Zhang and Yinhai Wang are with Civil and Environmental Engineering Department, University of Washington, Seattle, USA}
\thanks{Xinyu Zheng is with the College of Transportation, Tongji University, Shanghai, China}
\thanks{* Kehua Chen and Yinhai Wang are the corresponding authors, E-mail: zeonchen@uw.edu; yinhai@uw.edu.}
}

\markboth{Journal of \LaTeX\ Class Files,~Vol.~14, No.~8, August~2021}%
{Shell \MakeLowercase{\textit{et al.}}: A Sample Article Using IEEEtran.cls for IEEE Journals}


\maketitle

\begin{abstract}
Accurate pedestrian trajectory prediction is crucial for ensuring safety and efficiency in autonomous driving and human-robot interaction scenarios. Earlier studies primarily utilized sufficient observational data to predict future trajectories. However, in real-world scenarios, such as pedestrians suddenly emerging from blind spots, sufficient observational data is often unavailable (i.e. momentary trajectory), making accurate prediction challenging and increasing the risk of traffic accidents. Therefore, advancing research on pedestrian trajectory prediction under extreme scenarios is critical for enhancing traffic safety. In this work, we propose a novel framework termed \textit{Diffusion$^\mathbf{2}$}, tailored for momentary trajectory prediction. \textit{Diffusion$^\mathbf{2}$} consists of two sequentially connected diffusion models: one for \textit{backward prediction}, which generates unobserved historical trajectories, and the other for \textit{forward prediction}, which forecasts future trajectories. Given that the generated unobserved historical trajectories may introduce additional noise, we propose a dual-head parameterization mechanism to estimate their aleatoric uncertainty and design a temporally adaptive noise module that dynamically modulates the noise scale in the forward diffusion process. Empirically, \textit{Diffusion$^\mathbf{2}$} sets a new state-of-the-art in momentary trajectory prediction on ETH/UCY and Stanford Drone datasets.
\end{abstract}


\section{Introduction}
Accurate pedestrian trajectory prediction is essential for autonomous driving, as it enhances vehicle safety—particularly in scenarios involving human-vehicle interactions~\cite{zhang2024smartcooper, fang2024pacp}. In recent years, a wide variety of approaches have been developed to tackle this task. Typically, these approaches focus on forecasting future trajectories based on a sufficiently long observation period, such as 8 frames (about 3.2 seconds) \cite{yao2024trajclip, shi2023trajectory, wang2024trajfine}. 
However, in many real-world scenarios, such as when a pedestrian suddenly appears behind an obstacle (e.g., a building or a truck), the vehicle lacks sufficient time to gather adequate observational data. This poses significant challenges to conventional prediction methods, often leading to a substantial degradation in prediction performance \cite{sun2022human} and even increasing the risk of traffic accidents. 
As reported in~\cite{sun2022human}, the occurrence frequency of pedestrians with only momentary observations reaches $2.22\ s^{-1}$ in SDD~\cite{robicquet2016learning} and $1.02\ s^{-1}$ in ETH/UCY~\cite{pellegrini2009you} datasets. Therefore, it is crucial to investigate pedestrian trajectory prediction using limited observational data.

To address the aforementioned challenges, \cite{sun2022human} proposed \textit{momentary trajectory prediction} using only two observed frames. They introduced a Momentary Observation feature Extractor (MOE) that integrates social and scene context at the data level and improves representation learning via soft pre-training with masked trajectory and context restoration tasks.
\cite{monti2022many} proposed a teacher–student framework where the teacher, trained on rich data, guides a student with limited inputs. However, using only two frames restricts temporal understanding. To bridge this gap, \cite{li2023bcdiff} deployed a bidirectional consistent diffusion framework (BCDiff) to jointly generate both unobserved historical trajectories and future trajectories.  

\begin{figure*}
  \centering
  \includegraphics[width=0.85\linewidth]{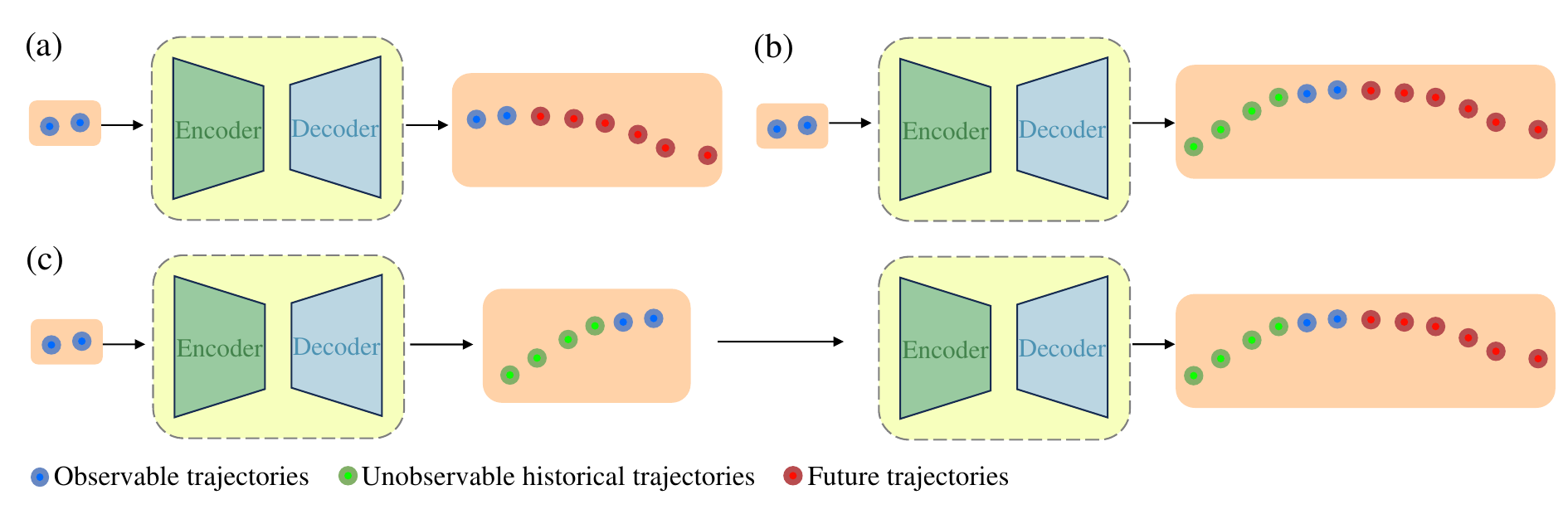}
  \caption{Three distinct frameworks are proposed to address momentary trajectory prediction. (a) The model directly utilizes observable trajectories to predict future trajectories. (b) The approach jointly predicts unobservable historical trajectories and future trajectories. (c) Our proposed framework, \textit{Diffusion$^\mathbf{2}$}, consists of two sequentially connected diffusion models: one dedicated to \textit{backward prediction}, and the other to \textit{forward prediction}.}
  \label{fig:DiffFrame}
\end{figure*}

While \cite{li2023bcdiff} has demonstrated the effectiveness of jointly predicting historical trajectories to enrich contextual information and compensate for the limitations of momentary observations, it involved simultaneously co-predicting historical and future trajectories, overlooking the causal relationship between the two components. In this study, we predict historical and future trajectories separately in a sequential manner. Specifically, we propose a model named \textit{Diffusion$^\mathbf{2}$} for momentary pedestrian trajectory prediction.
Building on the strengths of diffusion models, our framework employs two sequentially connected components: one for \textit{backward prediction} and the other for \textit{forward prediction}, as illustrated in Fig.~\ref{fig:DiffFrame}.
A follow-up concern is the reliability of the predicted historical trajectories: if they are highly inaccurate or noisy, they may adversely affect the subsequent prediction of future trajectories. Therefore, it is essential to quantify the uncertainty of the predicted historical trajectories and selectively utilize reliable information for downstream prediction tasks. However, diffusion models only implicitly learn data distributions and lack the ability to explicitly quantify uncertainty. To fill this gap, we design a dual-head parameterization mechanism that augments a standard noise prediction network by introducing two output heads. The first head predicts the noise, while the second estimates the log-variance for each coordinate, thereby enabling direct quantification of aleatoric uncertainty.

Afterwards, to incorporate the estimated aleatoric uncertainties, we design a temporally adaptive noise scheduler that dynamically injects noise into the forward diffusion model based on the predicted uncertainty levels. Intuitively, when historical trajectories are highly uncertain, more noise is injected into the forward diffusion model to encourage exploration during generation, and less noise is added when uncertainty is low.

To sum up, the main contributions of our study are:
\begin{itemize}
    \item We propose \textit{Diffusion$^\mathbf{2}$}, a novel framework consisting of two sequential diffusion models that capture the causal dependencies of trajectories: one for \textit{backward prediction} and the other for \textit{forward prediction}.
    \item We propose a dual-head parameterization mechanism that enables the diffusion model for \textit{backward prediction} to quantify aleatoric uncertainty in a single sampling pass, and further introduce an adaptive noise scheduling strategy for the \textit{forward prediction} diffusion model, which dynamically adjusts the noise magnitude based on the estimated uncertainty.  
    \item Our model achieves state-of-the-art accuracy in momentary pedestrian prediction, as demonstrated on the ETH/UCY and Stanford Drone datasets.
\end{itemize}

\section{Related Works}
\subsection{Traditional Trajectory Prediction}
Traditional human trajectory prediction tasks focus on forecasting future movements based on sufficiently long historical observations. To enhance the prediction accuracy, a variety of methods have been proposed to capture the temporal features and spatial features, with a particular focus on inter-pedestrian interactions and pedestrian-environment relationships. To capture the temporal features, researchers typically employed Long-Short Term Memory (LSTM) \cite{huang2019stgat, Alahi_2016_CVPR, salzmann2020trajectron++, luo2024granp, long2025physical} and Recurrent Neural Networks (RNNs) \cite{xu2022socialvae, moon2019prediction, Adeli_2021_ICCV} to model the temporal dynamics of pedestrian states. For modeling spatial interactions, social pooling mechanism \cite{alahi2016social, wong2024socialcircle}, graph-based modeling \cite{Mohamed_2020_CVPR, peng2021stirnet, sheng2024ego, tian2025risk} and attention mechanism \cite{mangalam2020not, sadeghian2019sophie, tsao2022social, sheng2024kinematics} are commonly employed to represent inter-agent and agent-environment dynamics. In addition, many physics-based models~\cite{tian2025mean}, such as Social Force Model (SFM) \cite{helbing1995social, luber2010people, zhang2023forceformer} and Energy-based Models \cite{wang2022seem, xiang2024socialcvae}, have been utilized to provide more interactive information. However, the aforementioned methods assume ideal conditions with ample historical observations, and therefore may not generalize well to extreme cases \cite{sun2022human}. In contrast, this work focuses on trajectory prediction under extreme scenarios, aiming to accurately forecast future trajectories given only two observed frames.

\subsection{Momentary Trajectory Prediction}
Momentary trajectory prediction aims to explore the limits of forecasting accuracy under extreme conditions, where only two historical frames are provided as input. In \cite{sun2022human}, the authors proposed a novel fusion mechanism that integrates social and scene context information at the data level, and then pre-trained the Momentary Observation Feature Extraction (MOE) module on several sub-tasks, such as masked trajectory completion and context restoration, to learn joint representations. To address the lack of temporal information under limited observations, \cite{monti2022many} adopted a teacher-student framework, where the teacher network, trained on rich 8-frame sequences, distills its spatial-temporal reasoning ability into a student network that learns to make accurate predictions using only 2 frames. As an alternative approach, \cite{li2023bcdiff} proposed a Bi-directional Consistent Diffusion framework (BCDiff) that jointly infers unobserved historical and future trajectories in a step-by-step manner to capture the bidirectional dependencies between them.

\subsection{Diffusion Models}
Diffusion models have achieved remarkable success in a variety of challenging domains \cite{chen2025dynamic, chen2025score, peng2024diffusion}, such as image synthesis \cite{ho2020denoising, song2019generative, rombach2022high}, computer vision \cite{cai2020learning, kawar2021stochastic, saharia2022image}, and natural language processing \cite{gong2022diffuseq, wu2023ar}, demonstrating their versatility and robustness. Recently, diffusion models have been deployed to trajectory prediction task. MID \cite{gu2022stochastic} designed a transformer-based network architecture to model the complex temporal dependencies in trajectories. EquiDiff \cite{chen2023equidiff} integrated diffusion modeling with equivariant transformers to ensure equivariant properties throughout the prediction process. LED \cite{mao2023leapfrog} employed a trainable leapfrog initializer in the trajectory prediction task to skip a large number of denoising steps, thereby significantly reducing inference time. In our work, we also leverage a diffusion model to achieve momentary trajectory prediction.

\subsection{Aleatoric Uncertainty}
Aleatoric uncertainty (AU) captures the irreducible, data-driven randomness present even when a model is perfectly specified. In trajectory prediction tasks, AU originates from sensor noise, measurement errors and the inherently stochastic nature of human or agent behaviour \cite{li2024unravelling}. Since this variability is tied to the data-collection process itself, AU cannot be reduced by collecting additional training data. The approaches to quantifying AU fall into two categories: explicit likelihood models and implicit likelihood models. Explicit likelihood models (e.g. heteroscedastic Gaussian \cite{rodrigues2018heteroscedastic} and neural processes \cite{garnelo2018neural}) output the full predictive distribution parameters directly; AU is then available analytically as the predictive variance. Implicit likelihood models (e.g. diffusion models \cite{ho2020denoising}, VAE \cite{kingma2013auto}) cannot provide a closed-form predictive variance, so AU must be approximated via Monte Carlo sampling. In this work, we propose a diffusion-based aleatoric uncertainty estimation method to quantify AU.

\section{Preliminaries}
\subsection{Problem Definition}
In this study, we aim to perform accurate momentary pedestrian trajectory prediction using only two observed frames as input. These observed frames are defined as 
$\mathbf{x}^{obs} = \left\{x_t, \mathbf{p}_t, \mathbf{S} \mid t = -1, 0\right\},$
where $x_t$ denotes the 2D position of the ego pedestrian, $\mathbf{p}_t \in \mathbb{R}^{N\times2}$ represents the positions of surrounding pedestrians, with $N$ denoting the number of surrounding pedestrians, and $\mathbf{S}$ corresponds to the semantic scene map. The goal is to predict future trajectories
$\mathbf{y} = \left\{x_t\right\}_{t=1}^{T_{fut}},$
with $T_{fut}$ indicating the prediction horizon. In addition, we define the unobserved historical trajectories as
$\mathbf{x}^{ubs} = \left\{x_t\right\}_{t=-1-T_{ubs}}^{-2},$
where $T_{ubs}$ is the length of the unobserved historical. Our model predicts future trajectories by leveraging inferred historical trajectories based on only a few observed points. Since the inferred historical trajectories may be noisy, our model also accounts for their uncertainty.

\subsection{Denoising Diffusion Probabilistic Model}
The Denoising Diffusion Probabilistic Model (DDPM) \cite{ho2020denoising} typically consists of two Markov chain processes: a forward diffusion process and a reverse denoising process. To avoid redundancy, we present the diffusion formulation for $\mathbf{y}$ only and note that the same process can be directly applied to $\mathbf{x}^{ubs}$. The forward diffusion process gradually corrupts the ground-truth data by adding Gaussian noise over multiple time steps, forming a sequence of perturbed samples $\left\{\mathbf{y}_0, \mathbf{y}_1, \ldots, \mathbf{y}_M\right\}$, where $\mathbf{y}_0$ is the original data and $M$ is the number of diffusion steps. Generally, we define the diffusion process as: 
\begin{align}
q(\mathbf{y}_{1:M} \mid \mathbf{y}_0) &:= \prod_{m=1}^{M} q(\mathbf{y}_m \mid \mathbf{y}_{n}), \\
q(\mathbf{y}_m \mid \mathbf{y}_0) &:= \mathcal{N}\!\big(\mathbf{y}_m;\alpha_m \mathbf{y}_0,\; \sigma_m^2\, \bm{I}\big).
\end{align}
where $q(\mathbf{y}_{M}) \sim \mathcal{N}(\mathbf{0}, \bm{I})$, $n=m-1$, and $\sigma_m$ denotes the standard deviation (so the covariance is $\sigma_m^2\bm{I})$.  The diffusion parameters $\alpha_m,\ \sigma_m > 0$ induce a noise schedule over the time steps such that the Signal-to-Noise Ratio (SNR) $\nu(m)=\alpha_m^{2}/\sigma_m^{2}$ is strictly monotonically decreasing in $m$.

Conversely, the reverse denoising process is trained to iteratively remove noise in a step-by-step fashion along this Markov chain, reconstructing the original trajectory conditioned on context information. Generally, we define the reverse denoising process as:
\begin{align}
p_{\theta}(\mathbf{y}_{0:M})
  &:= p(\mathbf{y}_M)\prod_{m=1}^{M} p_{\theta}(\mathbf{y}_{n}\mid \mathbf{y}_m),\\
p_{\theta}(\mathbf{y}_{n}\mid \mathbf{y}_{m})
  &:= \mathcal{N}\!\big(\boldsymbol{\mu}_{p},\,\boldsymbol{\Sigma}_{p}\big),\\
\boldsymbol{\mu}_{p}
  &= \frac{\alpha_{m\mid n}\,\sigma_{n}^{2}}{\sigma_{m}^{2}}\,\mathbf{y}_{m}
   + \frac{\sigma_{m\mid n}^{2}\,\alpha_{n}}{\sigma_{m}^{2}}\,
     \mathbf{y}_{\theta}(\mathbf{y}_{m},m),\\
\boldsymbol{\Sigma}_{p}
  &= \frac{\sigma_{n}^{2}\,\sigma_{m\mid n}^{2}}{\sigma_{m}^{2}}\,\bm{I}.
\label{eq:ddpm_p_theta}
\end{align}
where $\alpha_{m\mid n} = \frac{\alpha_m}{\alpha_n}$, and $\sigma^{2}_{m\mid n} = \sigma_m^{2} - {\alpha^{2}_{m\mid n}}{\sigma_n^{2}}$; $\mathbf{y}_{\theta}$ denotes the neural network with parameters $\theta$ to approximate $\mathbf{y}_{0}$, and $p(\mathbf{y}_M)$ is the standard Gaussian noise used as the starting point. In addition, $\boldsymbol{\mu}_{p}$ and $\boldsymbol{\Sigma}_{p}$ denote the predicted mean and variance, respectively.

\section{Methods}

\subsection{Overall Architecture}
\begin{figure*}
  \centering
  \includegraphics[width=1\linewidth]{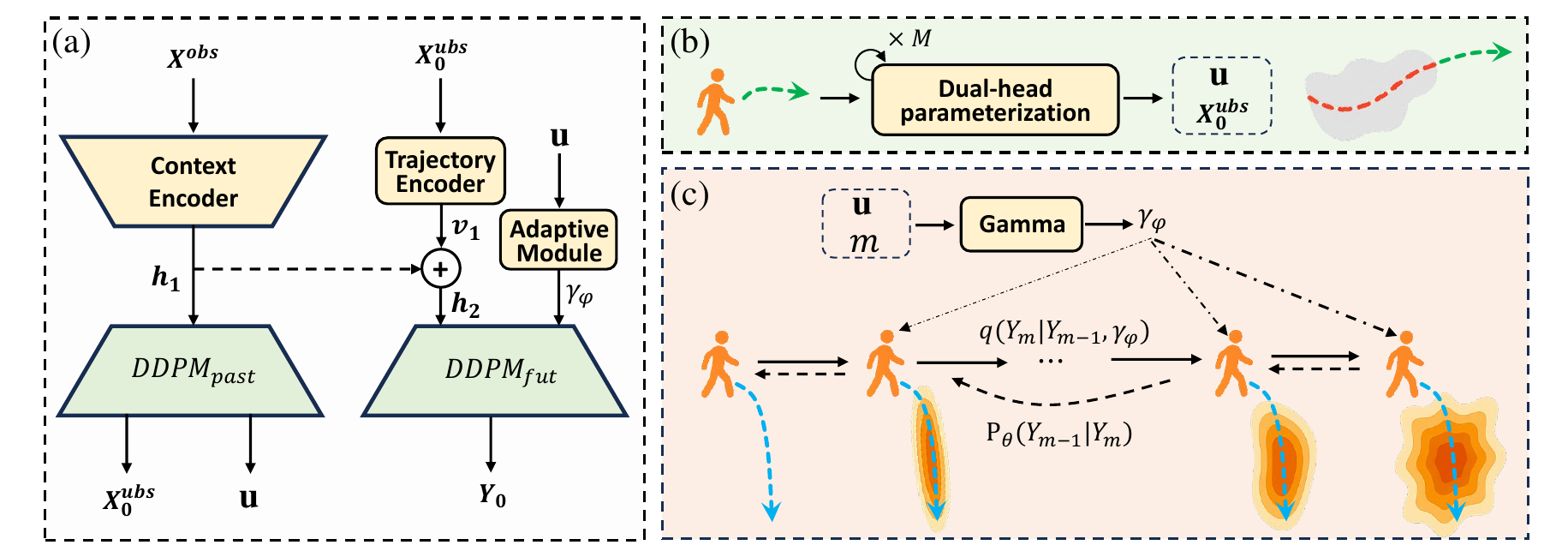}
  \caption{The overview of our proposed \textit{Diffusion$^\mathbf{2}$}:
(a) \textbf{Framework}: \textit{Diffusion$^\mathbf{2}$} consists of two sequentially connected diffusion models, $\textit{DDPM}_{past}$ and $\textit{DDPM}_{fut}$. $\textit{DDPM}_{past}$ simultaneously reconstructs the unobserved historical trajectory $\mathbf{x}^{ubs}_0$ and estimates its associated uncertainty $\mathbf{u}$. Then, $\textit{DDPM}_{fut}$ predicts the future trajectories $\mathbf{y}_0$. Both diffusion models share a common encoder that extracts contextual features $\mathbf{h_1}$ from the observed history $\mathbf{x}^{obs}$, while $\textit{DDPM}_{fut}$ employs an additional trajectory encoder to extract features $\mathbf{v}_1$ from the predicted unobserved historical trajectory $\mathbf{x}^{ubs}_0$. (b) \textbf{Dual-head Parameterization Mechanism}: The standard noise prediction network is augmented with two output heads, one for predicting noise and the other for predicting variance, which corresponds to aleatoric uncertainty.
(c) \textbf{Learnable Temporally Adaptive Noise Scheduling}: A gamma module takes $\mathbf{u}$ and the diffusion step $m$ as inputs and dynamically modulates the noise scale in the forward diffusion process of $\textit{DDPM}_{fut}$. Note that the red dashed line indicates the unobserved historical trajectory, with gray shading representing its uncertainty, while the green and blue dashed lines denote the observed history and future trajectories, respectively.
  }
  \label{fig:framework}
\end{figure*}

Our proposed model, \textit{Diffusion$^\mathbf{2}$}, is a two-stage diffusion framework consisting of two sequentially connected modules: $\textit{DDPM}_{past}$, which performs backward trajectory prediction and estimates aleatoric uncertainty, and $\textit{DDPM}_{fut}$, which predicts future trajectories, as illustrated in Fig.~\ref{fig:framework}. Given the observed frames $\mathbf{x}^{obs}$, a shared encoder first extracts a contextual representation $\mathbf{h}_1$. Conditioned on $\mathbf{h}_1$, $\textit{DDPM}_{past}$ predicts the unobserved historical trajectory $\mathbf{x}^{ubs}_0$ and simultaneously estimates its aleatoric uncertainty $\mathbf{u}$ through a dual-head parameterization mechanism. Sequentially, we employ a trajectory encoder consisting of two LSTM layers followed by a three-layers MLP to extract the features $\mathbf{v}_1$ from $\mathbf{x}^{ubs}_0$. The extracted feature $\mathbf{v}_1$ is then fused with the contextual representation $\mathbf{h}_1$ to form a new conditioning vector $\mathbf{h}_2$, which serves as the input context for $\textit{DDPM}_{fut}$ to predict future trajectories $\mathbf{y}_0$.
Furthermore, to incorporate the estimated aleatoric uncertainties, we introduce a learned temporally adaptive noise scheduling strategy for $\textit{DDPM}_{fut}$, where a gamma module dynamically adjusts the noise scale based on $\mathbf{u}$ and the current diffusion timestep $m$. Note that designing a sophisticated network to extract spatio-temporal features from trajectories is not the main focus of this work, and \textit{Diffusion$^\mathbf{2}$} is an encoder-agnostic framework that can seamlessly plug into the diverse encoders proposed in prior studies. In our experiments, we adopt the MOE \cite{sun2022human} encoder owing to its superior representational capacity. As the denoising backbone for both $\textit{DDPM}_{past}$ and $\textit{DDPM}_{fut}$, we adopt a streamlined Transformer decoder \cite{gu2022stochastic}. Parallel fully connected layers first project the corrupted trajectory sample and its contextual features into a common latent space, after which we add a sinusoidal encoding of the current diffusion timestep. This sequence is then processed by three stacked self-attention blocks that capture spatio-temporal dependencies. A final linear layer subsequently maps the representation back to 2-D coordinate space.

\subsection{Dual-head Parameterization}
In this work, to quantify the aleatoric uncertainty of predicted unobservable historical trajectories efficiently, we design a dual-head parameterization mechanism that augment the standard noise prediction network to output two heads:
\begin{equation}
    (\epsilon_{\theta_1}(\mathbf{x}^{ubs}_m, m, \mathbf{h}_1), \boldsymbol\ell_{\theta_1}(\mathbf{x}^{ubs}_m, m, \mathbf{h}_1)),
\end{equation}
where $\theta_1$ denotes the learned parameters of $\textit{DDPM}_{past}$. The first head, $\epsilon_{\theta_1}(\cdot)$, predicts the scaled Gaussian noise as in standard diffusion models, while the second head,  $\boldsymbol\ell_{\theta_1}(\cdot)$, directly predicts the log-variance on per-coordinate basis. For brevity, we abbreviate $\boldsymbol\ell_{\theta_1}(\mathbf{x}^{obs}_m, m, \mathbf{h}_1)$ by $\boldsymbol\ell_{m}$ and $\epsilon_{\theta_1}(\mathbf{x}^{obs}_m, m, \mathbf{h}_1)$ by ${\hat\epsilon}_{m}$ in what follows. During the inference phase, the dual-head parameterization propagates uncertainty without Monte-Carlo averaging. We update the reverse process as follows:
\begin{align}
&p_{\theta_1}\!\left(x^{\mathrm{ubs}}_{m-1}\mid x^{\mathrm{ubs}}_{m},\, m,\, \mathbf{h}_1\right)
= \mathcal{N}\!\big(\boldsymbol\mu_{\theta_1},\, \boldsymbol\Sigma_{\theta_1}\big), \label{eq:p_theta1}\\
&\boldsymbol\mu_{\theta_1}
= \frac{\alpha_{m\mid n}\,\sigma_n^{2}}{\sigma_m^{2}}\,\mathbf x^{\mathrm{ubs}}_{m}
  + \frac{\sigma_{m\mid n}^{2}\,\alpha_n}{\sigma_m^{2}}\,
    \frac{\mathbf x^{\mathrm{ubs}}_{m} - \sigma_m\,{\hat{\epsilon}}_m}{\alpha_m}, \label{eq:mu_theta1}\\
&\boldsymbol\Sigma_{\theta_1}
= \operatorname{diag}\!\big(\exp(\boldsymbol\ell_m)\big)
  + \frac{\sigma_n^{2}\,\sigma_{m\mid n}^{2}}{\sigma_m^{2}}\,\bm I, \label{eq:Sigma_theta1}
\end{align}
where $\operatorname{diag}(\cdot)$ denotes the operator that maps a vector to the diagonal matrix formed from its entries, with $\boldsymbol\mu_{\theta_1}$ and $\boldsymbol\Sigma_{\theta_1}$ denoting the predicted mean and covariance of $\textit{DDPM}_{past}$, respectively. Applying the Tweedie formula \cite{song2020score} yields a closed-form variance for the final prediction $\mathbf{x}^{ubs}_0$:
\begin{align}
\operatorname{diag}(\mathbf u)
&= \operatorname{diag}\!\big(\exp(\boldsymbol\ell_m)\big)
  + \frac{\sigma_n^{2}\,\sigma_{m\mid n}^{2}}{\sigma_m^{2}}\,\bm I.
\label{eq:u}
\end{align}
The first term captures the data-dependent aleatoric uncertainty, whereas the second term accounts for the known noise introduced by the scheduler. To isolate the intrinsic uncertainty, the second term can be subtracted. It is worth noting that the dual-head parameterization mechanism is designed for $\textit{DDPM}_{past}$.

\subsection{Temporally-Adaptive Learnable Noise Scheduling}
\label{Sec: LANS}
Conventional diffusion frameworks, with their preset forward noise schedules, often struggle to exploit known data features. To bridge this gap, we design a temporally-adaptive, uncertainty-aware noise schedule inspired by~\cite{sahoo2024mulan} that guides the diffusion process to better capture sample-specific structures and improves data distribution modeling. In particular, we leverage the uncertainty of predicted historical trajectories to adapt the noise schedule.

Naturally, the forward process for such noise schedule is defined as:
\begin{align}
    q(\mathbf{y}_m \mid \mathbf{y}_0, \mathbf{u}) &:= \mathcal{N}\!\big(\mathbf{y}_m;\boldsymbol\alpha_m(\mathbf{u}) \mathbf{y}_0,\; \operatorname{diag}(\boldsymbol\sigma_m^2(\mathbf{u})\big).
\end{align}
where $\mathbf{y}_m,\mathbf{y}_0 \in \mathbb{R}^{T_{fur} \times 2}$ and $\boldsymbol{\alpha}_m(\mathbf{u}),\boldsymbol{\sigma}_m(\mathbf{u}) \in \mathbb{R}^{T_{fut}}_+$. Then, the denoising process is given by:
\begin{align}
&p_{\theta_2}(\mathbf{y}_{n}\mid \mathbf{y}_{m}, \mathbf{h}_2, m, \mathbf{u})
  := \mathcal{N}\!\big(\boldsymbol{\mu}_{\theta_2},\,\boldsymbol{\Sigma}_{\theta_2}\big), \label{eq:p_theta_2}\\
&\boldsymbol{\mu}_{\theta_2}
  = \frac{\boldsymbol\alpha_{m\mid n}(\mathbf{u})\,\boldsymbol\sigma_{n}^{2}(\mathbf{u})}{\boldsymbol\sigma_{m}^{2}(\mathbf{u})}\,\mathbf{y}_{m}
   + \frac{\boldsymbol\sigma_{m\mid n}^{2}(\mathbf{u})\,\boldsymbol\alpha_{n}(\mathbf{u})}{\boldsymbol\sigma_{m}^{2}(\mathbf{u})}\,
     \mathbf{y}_{\theta}(\mathbf{y}_{m}, \mathbf{h}_2, m, \mathbf{u}), \label{eq:mu_theta_2}\\
&\boldsymbol{\Sigma}_{\theta_2}
  = \frac{\boldsymbol\sigma_{n}^{2}(\mathbf{u})\,\boldsymbol\sigma_{m\mid n}^{2}(\mathbf{u})}{\boldsymbol\sigma_{m}^{2}(\mathbf{u})}\,\bm{I}. \label{eq:sigma_theta_2}
\end{align}
where $\theta_2$ denotes the parameters of $\textit{DDPM}_{fut}$. Following \cite{sahoo2024mulan}, we control the the forward dynamic through a bounded log-SNR field $\boldsymbol\gamma_{\varphi}(\mathbf{u},m)$. Given $\boldsymbol\gamma_{\varphi}(\mathbf{u},m)$, the diffusion parameters are recovered as $\boldsymbol\alpha_m^{2}(\mathbf{u}) = \operatorname{sigmoid}\!\big(-\boldsymbol\gamma_{\varphi}(\mathbf{u},m)\big)$, $\boldsymbol\sigma_m^{2}(\mathbf{c}) = \operatorname{sigmoid}\!\big(\boldsymbol\gamma_{\varphi}(\mathbf{u},m)\big)$, and $\boldsymbol\nu(\mathbf{u},m) = \exp\!\big(-\boldsymbol\gamma_{\varphi}(\mathbf{u},m)\big)$. Here, $\boldsymbol\gamma_{\varphi}(\mathbf{u},m)$ is defined as: 
\begin{equation}
    \boldsymbol\gamma_{\varphi}(\mathbf{u}, m) = \gamma_\text{min} + (\gamma_\text{max} - \gamma_\text{min})\frac{f_{\varphi}(\mathbf{u}, m)}{f_{\varphi}(\mathbf{u},\, m=M)}
    \label{eq: gamma_function}
\end{equation}
where $f_{\varphi}(\mathbf{u}, m)$ is a  monotonic degree 5 polynomial in $m$:
\begin{align}
&f_\varphi(\mathbf{u}, m)
= \frac{a_1^2(\mathbf{u})}{5} m^5
 + \frac{a_1(\mathbf{u}) a_2(\mathbf{u})}{2} m^4 \nonumber\\
&+ \frac{a_2^2(\mathbf{u}) + 2 a_1(\mathbf{u}) a_3(\mathbf{u})}{3} m^3
 + a_2(\mathbf{u}) a_3(\mathbf{u}) m^2
 + a_3^2(\mathbf{u}) m + \mathcal{C}.  
\end{align}
This parameterization enforces the boundary conditions
$\boldsymbol\gamma_{\varphi}(\mathbf{u},0)=\gamma_{\min}$ and
$\boldsymbol\gamma_{\varphi}(\mathbf{u},M)=\gamma_{\max}$.
We implement the coefficient predictor as a neural network comprising two LSTM layers
followed by a two-layer MLP. The network outputs three coefficient vectors
$a_1, a_2,a_3 \in \mathbb{R}^{T_{fut}}$.
Following \cite{kingma2021variational,zheng2023improved}, we set $\gamma_{\min}=-13.30$ and $\gamma_{\max}=5.0$. Here, $\mathcal{C}$ denotes a constant.

\subsection{Loss Function} 

The loss function of DDPM is derived from the variational lower bound (VLB) of the negative log-likelihood of the data. \textit{Diffusion$^\mathbf{2}$} in this work comprises two sequentially connected diffusion models. Thus, the overall loss function naturally decomposes into two components, each corresponding to one direction of prediction. 

\noindent\textbf{Loss for $\textit{DDPM}_{past}$.} For \textit{backward prediction}, we design a dual-head parameterization mechanism that enables the diffusion model to estimate both the mean and aleatoric uncertainty. Concretely, the one-step reverse transition in the past branch is specified as Eq.~\ref{eq:p_theta1}, with the $\epsilon$-parameterized mean and diagonal covariance given by Eqs.~\ref{eq:mu_theta1} and \ref{eq:Sigma_theta1}. Naturally, the evidence lower bound (ELBO) contribution at step $m$ is the Gaussian negative log-likelihood:
\begin{equation}
\mathcal{L}^{\text{NLL}}_m
=\tfrac12\!\left(\mathbf{x}^{\mathrm{ubs}}_{m-1}-\boldsymbol{\mu}_{\theta_1}\right)^{\!\top}
\boldsymbol{\Sigma}_{\theta_1}^{-1}
\left(\mathbf{x}^{\mathrm{ubs}}_{m-1}-\boldsymbol{\mu}_{\theta_1}\right)
+\tfrac12\log\!\left|\boldsymbol{\Sigma}_{\theta_1}\right| + \mathcal{C},
\label{eq:nll-x}
\end{equation}
where $\mathcal{C}$ also collects terms independent of the learnable parameters. Replacing ${\hat\epsilon}_m$ in Eq.~\ref{eq:mu_theta1} with the true noise $\epsilon$ defines the posterior mean $\boldsymbol{\mu}^\star$, and their difference is:
\begin{equation}
\boldsymbol{\mu}^{\star}-\boldsymbol{\mu}_{\theta_1}
= -\,\frac{\sigma_{m\mid n}^2\sigma_n\,}{\alpha_m\,\sigma_m}\,
\big(\epsilon-{\hat{\epsilon}}_m\big).
\label{eq:diff-mean}
\end{equation}
Using \eqref{eq:diff-mean} in the quadratic form of \eqref{eq:nll-x} and the diagonal structure of
$\boldsymbol{\Sigma}_{\theta_1}$, all schedule-only factors collapse into constants, leaving the
parameter-dependent precision weighting $\exp(-\boldsymbol{\ell}_m)$ on the $\epsilon$-residual. The
log-determinant simplifies analogously since $\boldsymbol{\Sigma}_{\theta_1}$ is diagonal. Discarding
terms independent of $\epsilon_{\theta_1}$ and $\boldsymbol{\ell}_m$ yields the heteroscedastic Gaussian negative log-likelihood we optimize:
\begin{equation}
\mathcal{L}_1
= \frac{1}{2}\,\exp(-\boldsymbol{\ell}_{m})\,
\big\|{\epsilon}-\epsilon_{\theta_1}(\mathbf{x}^{\mathrm{ubs}}_{m},\,m,\,\mathbf{h}_1)\big\|_2^2
\;+\; \frac{1}{2}\,\boldsymbol{\ell}_{m}\,.
\label{eq:L1}
\end{equation}
The first term performs precision-weighted regression in $\epsilon$-space, while the second is the
$\log|\boldsymbol{\Sigma}_{\theta_1}|$ regularizer that prevents vanishing-variance (overconfident) solutions. 

\noindent\textbf{Loss for $\textit{DDPM}_{fut}$.} For forward prediction,  we parameterize the one-step conditional as in Eq.~\ref{eq:p_theta_2}, with
$\boldsymbol{\mu}_{\theta_2}$ and $\boldsymbol{\Sigma}_{\theta_2}$ given by
Eqs.~\ref{eq:mu_theta_2} and \ref{eq:sigma_theta_2}. Under our context-conditioned Gaussian
transition definition, the forward process is
\begin{equation}
q(\mathbf{y}_m \mid \mathbf{y}_n, \mathbf{u})
:= \mathcal{N}\!\big(\boldsymbol\alpha_{m\mid n}(\mathbf{u})\,\mathbf{y}_n,\;\boldsymbol\sigma^2_{m\mid n}(\mathbf{u})\,\bm I\big).
\end{equation}
The corresponding conditional distribution of $\mathbf{y}_n$ given $\mathbf{y}_m$ and the clean target
$\mathbf{y}_0$ is
\begin{align}
&q(\mathbf{y}_n \mid \mathbf{y}_m, \mathbf{y}_0, \mathbf{u})
:= \mathcal{N}\!\big(\boldsymbol{\mu}_q,\;\boldsymbol{\Sigma}_q\big),\\
&\boldsymbol{\mu}_q
= \frac{\boldsymbol\alpha_{m\mid n}(\mathbf{u})\,\boldsymbol\sigma_{n}^{2}(\mathbf{u})}{\boldsymbol\sigma_{m}^{2}(\mathbf{u})}\,\mathbf{y}_{m}
 + \frac{\boldsymbol\sigma_{m\mid n}^{2}(\mathbf{u})\,\boldsymbol\alpha_{n}(\mathbf{u})}{\boldsymbol\sigma_{m}^{2}(\mathbf{u})}\,\mathbf{y}_{0},
\end{align}
where $\boldsymbol{\Sigma}_q$ equals Eq.~\ref{eq:sigma_theta_2}, i.e.,
$\boldsymbol{\Sigma}_q=\boldsymbol{\Sigma}_{\theta_2}
=\frac{\boldsymbol\sigma_n^2(\mathbf u)\,\boldsymbol\sigma_{m\mid n}^2(\mathbf u)}{\boldsymbol\sigma_m^2(\mathbf u)}\,\bm I$.

With matching covariances, the ELBO term reduces to the KL between two Gaussians that differ only in
their means:
\begin{align}
&D_{\mathrm{KL}}\!\big(q(\mathbf{y}_n\!\mid\!\mathbf{y}_m,\mathbf{y}_0,\mathbf u)\,\big\|\,p_{\theta_2}(\mathbf{y}_n\!\mid\!\mathbf{y}_m,\mathbf h_2,m,\mathbf u)\big)\\
&=\tfrac12\!\left(\boldsymbol{\mu}_{\theta_2}-\boldsymbol{\mu}_q\right)^{\!\top}
\boldsymbol{\Sigma}_{\theta_2}^{-1}
\left(\boldsymbol{\mu}_{\theta_2}-\boldsymbol{\mu}_q\right)\nonumber\\
&=\tfrac12\!\left(\mathbf y_{0}-\mathbf y_{\theta_2}(\mathbf y_m,\mathbf h_2,m,\mathbf u)\right)^{\!\top}
\mathbf J_{0\mid m}(\mathbf u,m)\nonumber\\
&\quad\times \left(\mathbf y_{0}-\mathbf y_{\theta_2}(\mathbf y_m,\mathbf h_2,m,\mathbf u)\right),
\end{align}
because
\begin{align}
&\boldsymbol{\mu}_{\theta_2}-\boldsymbol{\mu}_q
= \frac{\boldsymbol\sigma_{m\mid n}^{2}(\mathbf u)\,\boldsymbol\alpha_{n}(\mathbf u)}{\boldsymbol\sigma_{m}^{2}(\mathbf u)}\,
\big(\mathbf y_{\theta_2}(\mathbf y_m,\mathbf h_2,m,\mathbf u)-\mathbf y_0\big),\\
&\qquad
\boldsymbol{\Sigma}_{\theta_2}^{-1}
=\frac{\boldsymbol\sigma_m^2(\mathbf u)}{\boldsymbol\sigma_n^2(\mathbf u)\,\boldsymbol\sigma_{m\mid n}^2(\mathbf u)}\,\mathbf I,
\end{align}
and we define the (context-conditioned) posterior precision as follows:
\begin{align}
    \mathbf J_{0\mid m}(\mathbf u,m)
    &:= \frac{\boldsymbol\alpha_n(\mathbf u)^{2}\,\boldsymbol\sigma_{m\mid n}(\mathbf u)^{2}}{\boldsymbol\sigma_n(\mathbf u)^{2}\,\boldsymbol\sigma_m(\mathbf u)^{2}}\;\bm I \nonumber\\
    &\;\equiv\; \operatorname{diag}\!\big(\boldsymbol\nu(\mathbf u,n)-\boldsymbol\nu(\mathbf u,m)\big).
\end{align}
Dropping constants that are independent of $\theta_2$ and taking the expectation over timesteps
(e.g., by uniformly sampling $m$) yields the training loss used for the forward branch:
\begin{align}
\mathcal{L}_2
=&\frac{1}{2}\,\mathbb{E}\Bigg[\sum_{m=1}^{M}
\big(\mathbf{y}_{0}-\mathbf{y}_{\theta_2}(\mathbf{y}_{m},\mathbf{h}_2,m,\mathbf{u})\big)^{\!\top}\label{eq:L2}\\
&\times \operatorname{diag}(\boldsymbol\nu(\mathbf{u},n)-\boldsymbol\nu(\mathbf{u}, m)) \nonumber \big(\mathbf{y}_{0}-\mathbf{y}_{\theta_2}(\mathbf{y}_{m},\mathbf{h}_2,m,\mathbf{u})\big)
\Bigg].
\end{align}
Thus, the total loss function $\mathcal{L} = \mathcal{L}_1 + \mathcal{L}_2$, and the training phase is illustrated in Algorithm~\ref{alg:bcdiff-train}.

\begin{algorithm}[t]
\caption{Pseudocode for \textit{Diffusion}$^\mathbf{2}$ Training Phase}
\label{alg:bcdiff-train}
{\setstretch{1.2}
\begin{algorithmic}[1]
\While{Model not converged}
  \State Sample frames $\big(\mathbf{x}^{obs},\, \mathbf{x}^{ubs},\, \mathbf{y})$ from dataset
  \State Sample $m \sim \mathrm{Uniform}(1, M)$
  \State Encode: $\mathbf{h}_1 \gets \mathbf{x}^{obs}$
  \State Calculate the loss $\mathcal{L}_1$ using Eq.~\ref{eq:L1}.
  \State Sample $\mathbf{x}^{ubs}_M \sim \mathcal{N}(\mathbf{0}, \bm{I})$
  \For {$m=\{M, \ldots, 1\}$}
  \State Sample ${\epsilon} \sim \mathcal{N}(\mathbf{0}, \bm{I})$
  \State Sample ${\epsilon'} \sim \mathcal{N}(\mathbf{0}, \bm{I})$
  \State $\mathbf{x}_{m-1}^{ubs} = \boldsymbol{\mu}_{\theta_1} + \exp(\frac{1}{2}\boldsymbol\ell_m)\odot \epsilon + \frac{\sigma_n\sigma_{m\mid n}}{\sigma_m}\epsilon'$
  \EndFor
  \State Calculate $\mathbf{u}$ using Eq.~\ref{eq:u}.
  \State Encode: $\mathbf{v}_1 \gets \mathbf{x}^{\text{ubs}}_0$, 
  \State Concatenate: $\mathbf{h}_2 \gets \mathbf{v}_1 \oplus \mathbf{h}_1 $
  \State Sample $m \sim \mathrm{Uniform}(1, M)$
  \State Calculate the loss $\mathcal{L}_2$ using Eq.~\ref{eq:L2}.
  \State Take gradient descent step on $\nabla (\mathcal{L}=\mathcal{L}_1+\mathcal{L}_2)$, and \Statex \hspace{1.3em} then update the model's parameters.
\EndWhile 
\end{algorithmic}}
\end{algorithm}


\subsection{Model inference} 
The preceding sections detailed the model architecture and training procedure. Once training is complete, we can use \textit{Diffusion}$^{2}$ to generate future pedestrian trajectories. In the first stage, $\textit{DDPM}_{\text{past}}$ performs denoising to obtain plausible unobserved trajectories from pure Gaussian noise $\mathbf{x}_M^{ubs} \sim \mathcal{N}(\mathbf{0}, \bm{I})$. The denoising process proceeds as follows:
\begin{align}
    \mathbf{x}_{m-1}^{ubs} = \boldsymbol{\mu}_{\theta_1} + \exp(\frac{1}{2}\boldsymbol\ell_m)\odot \epsilon + \frac{\sigma_n\sigma_{m\mid n}}{\sigma_m}\epsilon'
\end{align}
Where $\epsilon$ and $\epsilon' \sim \mathcal{N}(\mathbf{0}, \bm{I})$. Meanwhile, leveraging the Tweedie formula~\cite{song2020score}, we can obtain a closed-form variance for the final prediction $\mathbf{x}_{0}^{ubs}$, as shown in Eq.~\ref{eq:u}. In the second stage, we employ the $\textit{DDPM}_{\text{future}}$ denoising procedure to forecast plausible future trajectories from a pure Gaussian noise $\mathbf{y}_M \sim \mathcal{N}(\mathbf{0}, \bm{I})$. Concretely, starting from $\mathbf{y}_M$, we iterate as follows:
\begin{align}
    \mathbf{y}_{m-1} = \boldsymbol{\mu}_{\theta_2} + \frac{\boldsymbol\sigma_n(\mathbf{u})\boldsymbol\sigma_{m\mid n}(\mathbf{u})}{\boldsymbol\sigma_m(\mathbf{u})}\epsilon
\end{align}
where $\epsilon \sim \mathcal{N}(\mathbf{0}, \bm{I})$. To summarize, the inference procedure is presented in Algorithm~\ref{alg:bcdiff-sample}.

\begin{algorithm}[t]
\caption{Pseudocode for \textit{Diffusion}$^\mathbf{2}$ Sampling Phase}
\label{alg:bcdiff-sample}
{\setstretch{1.2}
\begin{algorithmic}[1]
\State \textbf{Input:} observed frames $\mathbf{x}^{obs}$
\State \textbf{Output:} Future trajectories $\mathbf{y}$
\State Sample $\mathbf{x}^{ubs}_M \sim \mathcal{N}(\mathbf{0}, \bm{I})$
\State Encode: $\mathbf{h}_1 \gets \mathbf{x}^{obs}$
\State Same as Alg.~\ref{alg:bcdiff-train}, Steps 6–13
\For {$m=\{M, \ldots, 1\}$}
\State Sample $\mathbf{y}_M \sim \mathcal{N}(\mathbf{0}, \bm{I})$
\State Sample ${\epsilon} \sim \mathcal{N}(\mathbf{0}, \bm{I})$
\State $\mathbf{y}_{m-1} = \boldsymbol{\mu}_{\theta_2} + \frac{\boldsymbol\sigma_n(\mathbf{u})\boldsymbol\sigma_{m\mid n}(\mathbf{u})}{\boldsymbol\sigma_m(\mathbf{u})}\epsilon$
\EndFor
\State $\hat{\mathbf{y}} = \mathbf{y}_0$
\end{algorithmic}}
\end{algorithm}

\section{Experiments}
\subsection{Datasets}
We evaluate the performance of our proposed model on the ETH/UCY \cite{pellegrini2009you} and Stanford Drone Dataset (SDD) \cite{robicquet2016learning}. The ETH/UCY dataset comprises five distinct scenes (ETH, HOTEL, UNIV, ZARA1, and ZARA2), containing trajectories of 1,536 pedestrians sampled at 2.5 Hz. For ETH/UCY, following prior works \cite{li2023bcdiff, sun2021three, gu2022stochastic}, we adopt the widely used leave-one-scene-out protocol, training the model on four scenes and testing it on the held-out scene. The SDD is a large-scale dataset that captures trajectories of pedestrians, cyclists, and vehicles from a bird’s-eye view on a university campus, providing diverse and complex interaction scenarios for trajectory prediction research.

\subsection{Metrics}
In line with previous research \cite{li2023bcdiff, mangalam2021goals}, we adopt Average Displacement Error (ADE) and Final Displacement Error (FDE) as evaluation metrics to demonstrate the effectiveness of our proposed method. Specifically, ADE measures the average Euclidean distance between the predicted trajectory and the ground-truth trajectory over all time steps, and is defined as:
\begin{equation}
    \text{ADE} = \frac{1}{T_{fut}} \sum_{t=1}^{T_{fut}} \left\| \hat{\mathbf{y}}_t - \mathbf{y}_t \right\|_2,
\end{equation}
where \( \hat{\mathbf{y}}_t \) and \( \mathbf{y}_t \) denote the predicted and ground-truth positions at time step \( t \), respectively, and \( T \) denotes the total length of the trajectory. FDE, on the other hand, measures only the final-step prediction error, and is computed as:
\begin{equation}
    \text{FDE} = \left\| \hat{\mathbf{y}}_{T_{fut}} - \mathbf{y}_{T_{fut}} \right\|_2.
\end{equation}

To ensure a fair comparison, following prior work \cite{li2023bcdiff, gu2022stochastic, sun2021three}, we generate 20 predicted future trajectories and report the final error as the minimum error across all predictions.

\subsection{Baselines}
To demonstrate the effectiveness of our model, we compare prediction results against both conventional trajectory-prediction baselines and models specifically designed for momentary trajectory prediction:
\begin{itemize}
    \item \textbf{Social GAN} \cite{gupta2018social}: Social GAN employs an encoder–decoder architecture with a social-pooling module to capture multi-agent interactions.
    \item \textbf{Next} \cite{liang2019peeking}: Next leverages an attention-based LSTM coupled with spatial activity localization to extract features from behavioral features and person–scene interactions.
    \item \textbf{Social-STGCNN} \cite{mohamed2020social}: Social-STGCNN proposes a spatio-temporal graph convolutional network that models pedestrians as graph nodes to capture social interactions.
    \item \textbf{Trajectron++} \cite{salzmann2020trajectron++}: Trajectron++ leverages a graph-structured recurrent network to predict the trajectories of a general number of diverse agents. 
    \item \textbf{SGCN} \cite{shi2021sgcn}: Pedestrian trajectories are modeled with a sparse graph convolutional approach that uses directed spatial graphs for interactions and directed temporal graphs for motion patterns.
    \item \textbf{SingularTrajectory} \cite{bae2024singulartrajectory}: A diffusion-based, universal trajectory predictor maps diverse motion tasks into a shared embedding space, enabling generalizable predictions across settings.
    \item \textbf{STT+DTO} \cite{monti2022many}: A knowledge-distillation framework in which a student learns future-trajectory prediction from limited observations by mimicking a teacher trained on full trajectories.
    \item \textbf{MID} \cite{gu2022stochastic}: MID proposed a novel transformer-based  network to model complex temporal dependency in trajectories.
    \item \textbf{EigenTrajectory} \cite{bae2023eigentrajectory}: EigenTrajectory uses low-rank spatio-temporal descriptors to compactly represent and forecast multi-modal pedestrian motion.
    \item \textbf{PCCSNet} \cite{sun2021three}: PCCSNet proposes a novel framework that formulates multimodal prediction into three steps.
    \item \textbf{SocialVAE} \cite{xu2022socialvae}: SocialVAE employs a temporally-aware variational autoencoder (VAE) to model the multimodality and uncertainty in human trajectory prediction.  
    \item \textbf{MOE} \cite{sun2022human}: MOE designs a Momentary Observation feature Extractor to extract the features in momentary observations.
    \item \textbf{BCDiff} \cite{li2023bcdiff}: BCDiff proposes a bidirectional diffusion model that jointly generates unobserved historical and future trajectories.
\end{itemize}

\subsection{Experimental Setup}
For the forward diffusion process of $\textit{DDPM}_{past}$, we employ a linear noise schedule, fixing the variance bounds to $\beta_{1}=1\times10^{-4}$ and $\beta_{M}=0.05$. Our encoder uses exactly the hyper-parameter configuration reported in \cite{sun2022human}. Separately, the denoising backbone is composed of three Transformer encoder blocks, each with a model dimension of 128, a 256-dimensional feed-forward sub-layer, and 4 self-attention heads.

\textit{Diffusion$^\mathbf{2}$} was implemented in PyTorch and trained on a server running Ubuntu 22.04, equipped with an Intel Xeon Platinum 8358P (2.60 GHz) CPU and a single NVIDIA A800-SXM4-80G GPU. Training was carried out for 100 epochs with a batch size of 256.

\subsection{Main Results}
\noindent\textbf{Comparison with State-of-the-Art Methods.}
\begin{table*}[ht]
\caption{Comparisons of various approaches on multiple datasets. The evaluation results are reported as ADE/FDE (m). The best result is highlighted in bold, and the runner-up is \underline{underlined}.}
\centering
\resizebox{\textwidth}{!}{%
\begin{tabular}{lccccccc}
\hline
Momentary & ETH & HOTEL & UNIV & ZARA1 & ZARA2 & AVG & SDD \\
\hline
SGAN \cite{gupta2018social} & 0.86/1.60 & 0.52/0.99 & 0.57/1.20 & 0.41/0.79 & 0.36/0.74 & 0.54/1.06 & 17.76/34.83 \\
Next \cite{liang2019peeking}& 0.77/1.81 & 0.32/0.62 & 0.61/1.31 & 0.39/0.82 & 0.34/0.76 & 0.49/1.06 & - \\
Social-STGCNN \cite{mohamed2020social} & 1.24/2.23 & 0.77/1.44 & 0.45/0.81 & 0.38/0.57 & 0.35/0.58 & 0.64/1.13 & 17.77/29.12 \\
Trajectron++ \cite{salzmann2020trajectron++} & 0.76/1.43 & 0.30/0.56 & 0.36/0.74 & 0.22/0.42 & 0.18/0.34 & 0.36/0.70 & 13.07/22.88 \\
SGCN \cite{shi2021sgcn}& 0.88/1.66 & 0.55/1.16 & 0.38/0.71 & 0.30/0.54 & 0.25/0.46 & 0.47/0.91 & 15.40/25.69 \\
SingularTrajectory \cite{bae2024singulartrajectory}& 0.45/0.67 & 0.18/0.29 & \textbf{0.24}/\underline{0.43} & 0.19/\textbf{0.33} & 0.17/0.28 & 0.25/0.40 & - \\
STT+DTO \cite{monti2022many}& 0.62/1.22 & 0.29/0.56 & 0.58/1.14 & 0.45/0.98 & 0.34/0.74 & 0.46/0.93 & - \\
MID \cite{gu2022stochastic}& 0.63/1.05 & 0.29/0.49 & 0.30/0.56 & 0.30/0.56 & 0.22/0.40 & 0.35/0.61 & - \\
EigenTrajectory \cite{bae2023eigentrajectory} & 0.46/0.76 & 0.17/0.28 & \underline{0.25}/0.44 & 0.19/\underline{0.35} & \underline{0.15}/\underline{0.27} & 0.25/0.42 & - \\ \hline \hline
\multirow{3}{*}{\makecell[l]{%
  PCCSNet \cite{sun2021three}\\
  \quad +MOE \cite{sun2022human}\\
  \quad +BCDiff \cite{li2023bcdiff}\\
}}
& 0.34/0.65 & 0.14/0.25 & 0.31/0.63 & 0.23/0.46 & 0.16/0.37 & 0.24/0.47 & 9.19/17.71 \\
& 0.31/0.57 & \underline{0.13}/0.21 & \underline{0.25}/0.53 & 0.20/0.41 & \textbf{0.14}/0.31 & \underline{0.20}/0.41 & 8.40/16.08 \\ 
& \underline{0.30}/\underline{0.56} & \underline{0.13}/\underline{0.20} & \underline{0.25}/0.52 & \underline{0.18}/0.37 & \textbf{0.14}/0.31 & \textbf{0.19}/\underline{0.39} & \underline{8.32}/15.87 \\ 
\hline

\multirow{3}{*}{\makecell[l]{%
  SocialVAE \cite{xu2022socialvae}\\
  \quad +MOE \cite{sun2022human}\\
  \quad +BCDiff \cite{li2023bcdiff}\\
}}
& 0.64/1.10 & 0.21/0.34 & 0.27/0.51 & 0.22/0.39 & 0.18/0.34 & 0.30/0.54 & 9.56/16.10\\
& 0.57/1.01 & 0.17/0.29 & 0.26/0.44 & 0.22/0.36 & 0.17/0.32 & 0.28/0.48 & 9.12/14.98\\
& 0.53/0.91 & 0.17/0.27 & \textbf{0.24}/\textbf{0.40} & 0.21/0.37 & 0.16/\textbf{0.26} & 0.26/0.44 & 9.05/\textbf{14.86}\\
\hline\hline
\textit{Diffusion$^\mathbf{2}$} (ours) & \textbf{0.29}/\textbf{0.45} & \textbf{0.10}/\textbf{0.13} & 0.26/0.47 & \textbf{0.17}/0.37 & \textbf{0.14}/\textbf{0.26} & \textbf{0.19}/\textbf{0.33} &\textbf{8.26}/\underline{14.87}\\
\hline
\end{tabular}
}
\label{tab:comparison}
\end{table*}
We validate our proposed method on multiple datasets and the overall performance is shown in Tab.~\ref{tab:comparison}. The experimental results clearly show that \textit{Diffusion$^\mathbf{2}$} outperforms the baseline models on most datasets. Specifically, the ADE and FDE of \textit{Diffusion$^\mathbf{2}$} are markedly lower than the current state-of-the-art (SOTA) results. For example, on HOTEL, our ADE (0.10) is 23.1\% lower than the SOTA value (0.13), and our FDE (0.13) is 35.0\% lower than the SOTA (0.20). On ETH, our FDE (0.45) shows a 19.6\% reduction compared to the SOTA (0.56). Averaged across ETH/UCY, we obtain an FDE of 0.33, which is 15.4\% below the SOTA value of 0.39. Our ADE is 0.19, matching the reported SOTA (0.19) and exhibiting a slight improvement on the order of $10^{-3}$.
On SDD, \textit{Diffusion$^\mathbf{2}$} achieves an ADE/FDE of 8.26/14.87. This sets a new best ADE, improving upon the prior lowest value (8.32) by about 0.7\%, and the FDE essentially ties the best reported number (14.86), with a difference of 0.01 (i.e., $<10^{-2}$). These results suggest that our uncertainty-aware dual-head parameterization and adaptive noise scheduling transfer well to aerial-view, heterogeneous multi-agent scenes.
Compared with previous methods, \textit{Diffusion$^\mathbf{2}$} first uses a dual-head parameterization mechanism to estimate the uncertainty of historical trajectory predictions. Then, the adaptive noise scheduling module dynamically adjusts the forward noise addition process according to the uncertainty. Through these modules, the framework encourages the extraction and learning of informative features, thereby improving trajectory generation quality. Although overall performance is competitive, our method trails the state of the art on datasets such as UNIV. A plausible explanation is that the UNIV scene contains richer, more intricate social interactions (e.g., higher crowd density and frequent group formations), as shown in Fig.~\ref{fig:visu}. These cases motivate further refinement of our framework for better adaptability to diverse scene dynamics.

\noindent\textbf{Learned Adaptive Noise Schedule.}
\begin{figure}
  \centering
  \includegraphics[width=1\linewidth]{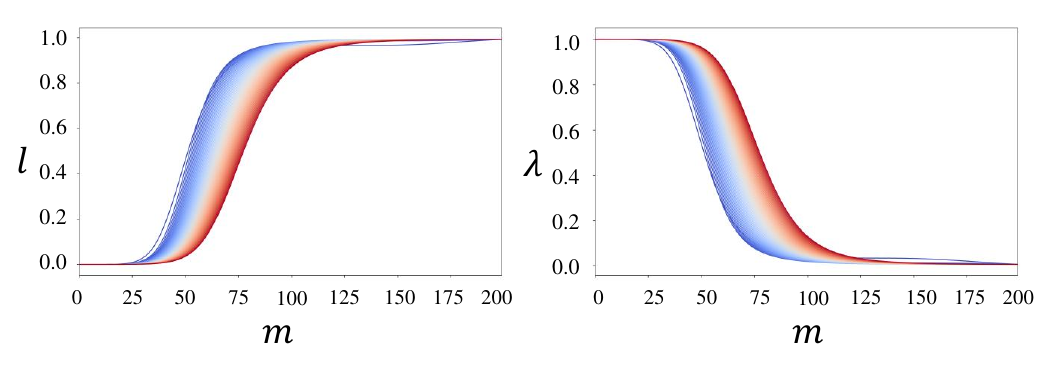}
  \caption{A visualization of the learned $\lambda$ and $l$ over diffusion step $m$, where each $l_m = \operatorname{sigmoid}(\gamma_{\varphi}(\mathbf{u}, m))$.}
  \label{fig:visu_sigma}
\end{figure}
Fig.~\ref{fig:visu_sigma} presents the learned schedules of the averaged \(\lambda_m\) and \(l_m\) over the diffusion steps \(m\). The colour gradient—from blue (low predictive uncertainty) to red (high predictive uncertainty)—highlights how the model adapts its noise‐injection strategy to the input’s uncertainty level. For low‐uncertainty inputs (blue curves), \(l_m\) rises steeply, so stronger noise is injected early in the forward process. Conversely, high‐uncertainty inputs (red curves) show a slower increase, delaying and attenuating the injected noise. The underlying reason for this attenuation is that highly uncertain inputs must preserve fragile signal features; by adding less noise, the model avoids drowning these features and enables the reverse process to recover them more reliably during reconstruction.

\noindent\textbf{Inference Latency.} 
In our proposed \textit{Diffusion$^\mathbf{2}$} framework, the DDPM-based sampler is a plug-and-play module that can be replaced by more efficient diffusion samplers without changing the rest of the architecture. We benchmark the inference latency of representative baselines and our model (with both DDPM and DDIM sampling) on identical hardware (NVIDIA RTX~A800). As summarized in Table~\ref{tab:inference_latency}, our DDPM-based instantiation runs at $412$\,ms, which is slower than discriminative baselines MID ($325$\,ms) and PCCSNet ($107$\,ms). This gap mainly comes from the multi-step Markovian denoising inherent to DDPM. Replacing the sampler with DDIM collapses the sampling chain to far fewer steps, reducing latency to $75$\,ms without changing the architecture, thus meeting real-time requirements. BCDiff~\cite{li2023bcdiff}, while also reconstructing unobserved history and predicting future trajectories like ours, is slow due to its iterative bidirectional refinement in the BCDU module (about $M{=}40$ steps), which incurs many forward passes and yields high latency ($\sim$4\,s). Meanwhile, many recent methods~\cite{mao2023leapfrog, zhang2022fast, shih2023parallel} have been proposed to accelerate inference by loosening Markovian constraints and generating trajectories with fewer or skipped steps, offering substantial potential to further reduce the inference latency of \textit{Diffusion$^\mathbf{2}$}.

\begin{figure*}
  \centering
  \includegraphics[width=0.9\linewidth]{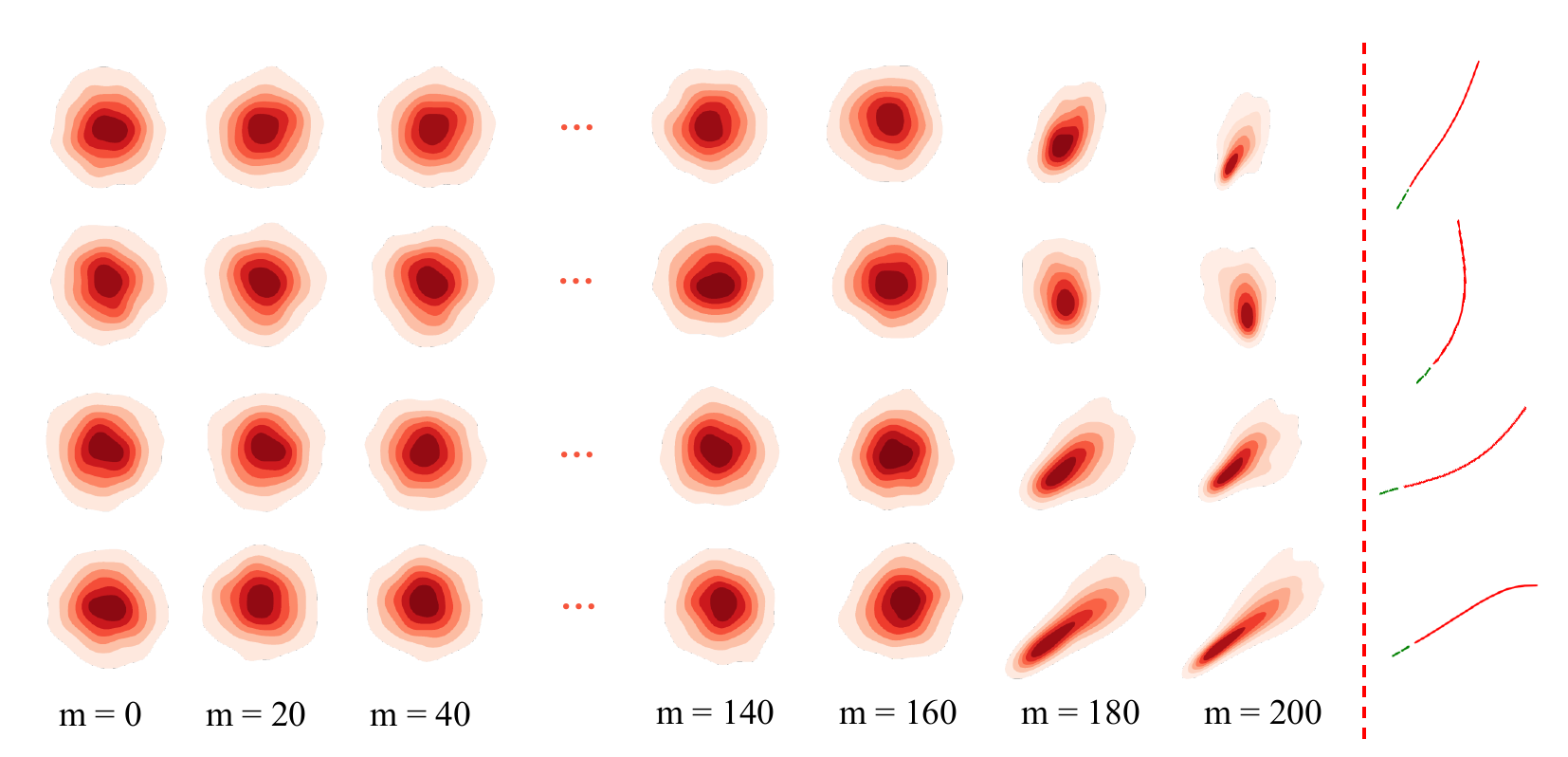}
  \caption{Visualization of generated trajectories across reverse-diffusion time $m$. The red and green lines represent ground-truth future and observable trajectories. As $m$ increases, the process progressively suppresses indeterminacy and converges toward the desired future path. Initialized from a pure gaussian noise at $m=0$ and conditioned on context information, the model gradually removes implausible hypotheses and aligns with the ground-truth future (red, solid).
  }
  \label{fig:denoising_process}
\end{figure*}

\noindent\textbf{Reverse Diffusion Process.}
As Fig. \ref{fig:denoising_process} illustrates, we visualize the denoising dynamics of the $\textit{DDPM}_{fut}$, rendering snapshots every 20 inference steps $(m \in \{0, 20, \dots, 200\})$. Starting from an initial Gaussian noise at $m=0$, the reverse process progressively suppresses uncertainty and converges toward a coherent future trajectory consistent with the observed history. Owing to the minimal distributional divergence observed during the initial 160 steps, intermediate samples for $m \in \{ 60,\dots,120\}$ are omitted to emphasize the salient transitions.

\begin{table}[t]
\centering
\small
\caption{Inference latency comparison on identical hardware (NVIDIA RTX~A800).}
\label{tab:inference_latency}
\setlength{\tabcolsep}{8pt}
\begin{tabular}{l r}
\toprule
\textbf{Method} & \textbf{Latency (ms)} \\
\midrule
MID~\cite{gu2022stochastic}        & 325 \\
PCCSNet~\cite{sun2021three}        & 107 \\
BCDiff~\cite{li2023bcdiff}         & $\sim$4000 \\
\textbf{Ours (DDPM)}               & 412 \\
\textbf{Ours (DDIM)}               & \textbf{75} \\
\bottomrule
\end{tabular}
\end{table}


\subsection{Qualitative evaluation.}
\begin{figure*}
  \centering
  \includegraphics[width=1\linewidth]{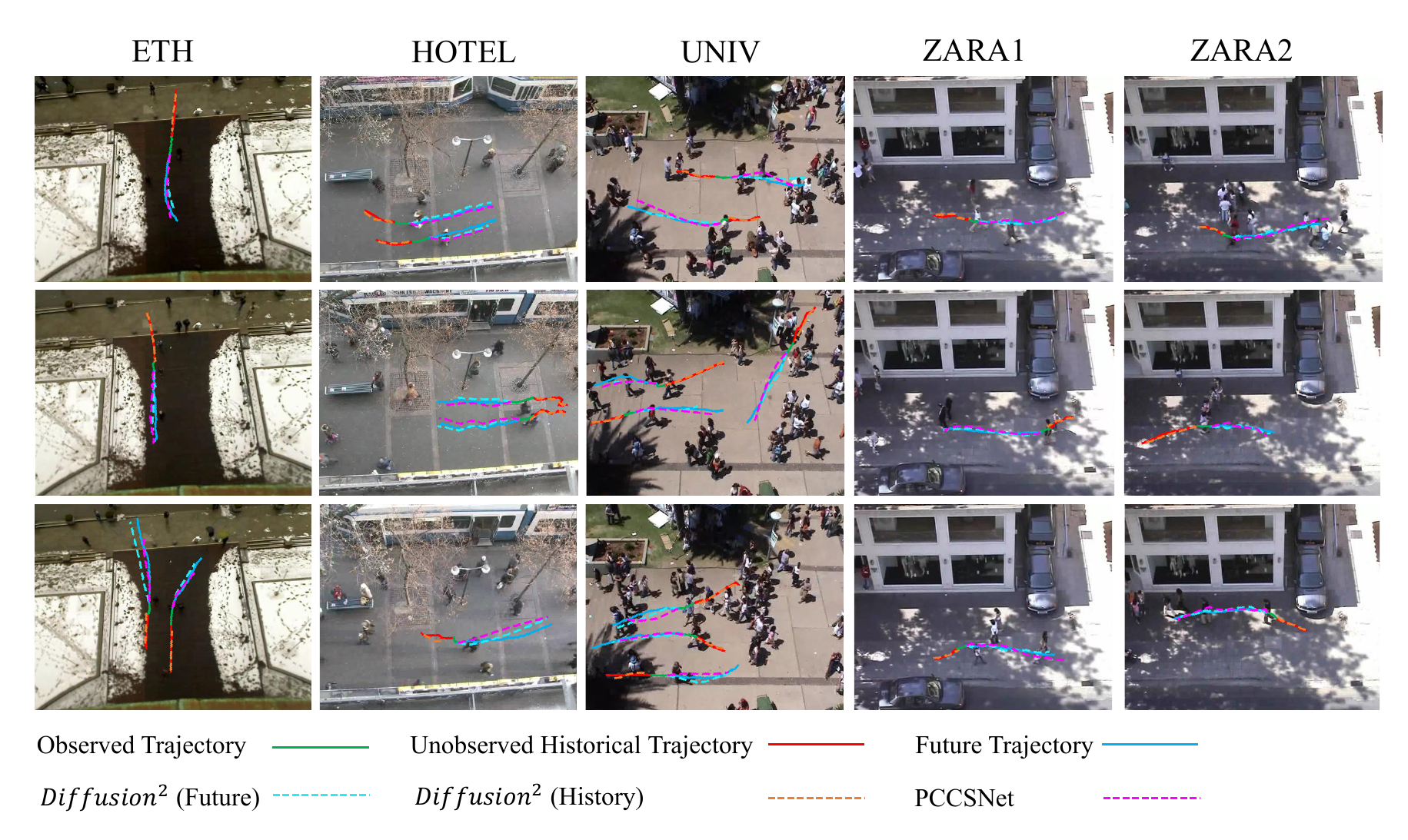}
  \caption{Visualization of predicted unobservable historical and future trajeactories on the ETH/UCY dataset. The red lines represent the unobserved historical trajectories, while the cyan lines denote the observed historical trajectories. The ground truth future trajectories are marked in light blue. Predictions from \textit{Diffusion$^\mathbf{2}$} are visualized as blue dashed lines for future and orange dashed lines for history, whereas the PCCSNet predictions are highlighted in magenta dashed lines.}
  \label{fig:visu}
\end{figure*}

\noindent\textbf{Qualitative Visualization.} We further demonstrate the effectiveness of our proposed method through qualitative results. Fig.~\ref{fig:visu} presents the predictions generated by our \textit{Diffusion$^\mathbf{2}$} framework and PCCSNet \cite{sun2021three} across all five scenes of the ETH/UCY dataset. In each scene, we visualize the observed trajectories, the unobserved historical paths, and the future ground truth movements. As shown, \textit{Diffusion$^\mathbf{2}$} produces predictions that closely follow the actual future and inferred past trajectories. PCCSNet also shows strong performance, particularly in short-term predictions. However, in some cases, PCCSNet exhibits slight deviations from the ground truth when predicting longer-term future paths. Overall, these results visually demonstrate the strong capability of \textit{Diffusion$^\mathbf{2}$} to model complex pedestrian dynamics under partial observability.

\begin{figure*}
  \centering
  \includegraphics[width=1\linewidth]{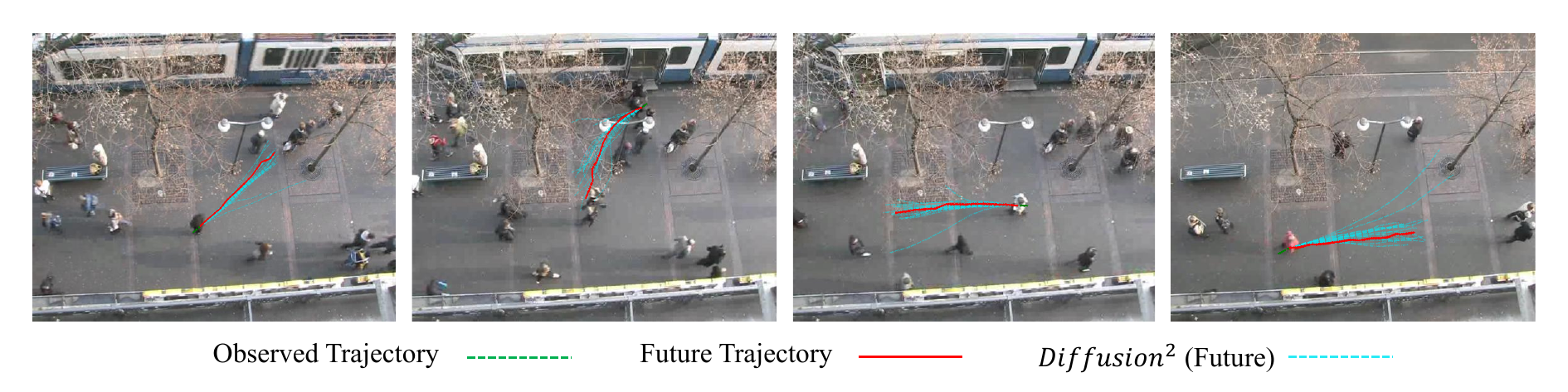}
  \caption{Visualization of predicted pedestrian trajectories across several scenes in the HOTEL dataset. Given the observed past (green, dashed), we show the ground-truth future (red, solid) and the model’s randomly sampled predictions (cyan, dashed). Trajectories are sampled from \textit{Diffusion}$^2$ and rendered after de-normalization to the scene frame. }
  \label{fig:multi}
\end{figure*}

\noindent\textbf{Multimodal.} As \textit{Diffusion}$^2$ learns an implicit distribution over future trajectories and is inherently multimodal, we visualize multiple predicted trajectories in Fig.~\ref{fig:multi}. As Fig.~\ref{fig:multi} shows, predicted trajectories from \textit{Diffusion}$^2$ align closely with the ground-truth future trajectories in most cases, while a minority of samples follow alternative directions, highlighting the model’s inherent multimodality.

\begin{figure*}
  \centering
  \includegraphics[width=1\linewidth]{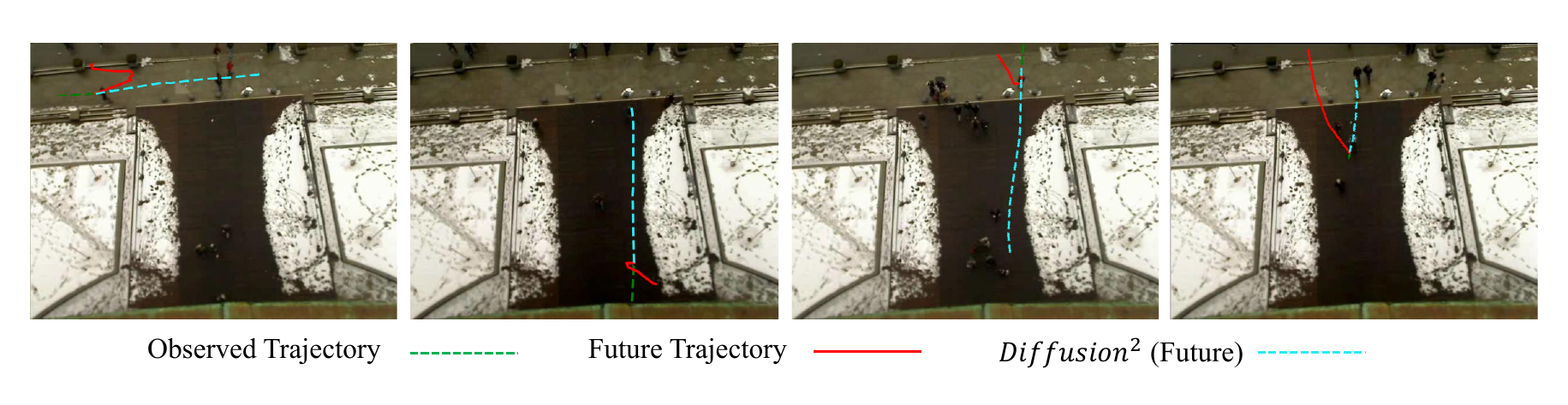}
  \caption{Visualization of predicted pedestrian trajectories across several scenes in ETH dataset (failure cases). Given the observed past (green, dashed), We show the ground-truth future (red, solid) and the model’s best-of-20 predicted trajectory. Trajectories are sampled from \textit{Diffusion}$^2$ and rendered after de-normalization to the scene frame. }
  \label{fig:fail}
\end{figure*}

\noindent\textbf{Failure Cases.} To probe when \textit{Diffusion}$^{2}$ fails, we examine the worst four cases on the ETH dataset (Fig.~\ref{fig:fail}), selected by best-of-20 ADE. We observe that errors spike when pedestrians undergo abrupt state changes, for example sudden heading reversals and sharp turns. In these moments the short observational window provides insufficient cues to anticipate the transition, leading \textit{Diffusion}$^{2}$ to extrapolate the pre-change motion and miss the new intent. 

\subsection{Ablation Studies}

\begin{table}[t]
  \centering
  \caption{Ablation study for our \textit{Diffusion$^\mathbf{2}$}.}
  \label{tab:mulan_uncertainty}
  \setlength{\tabcolsep}{3.8pt}
  \renewcommand{\arraystretch}{1.2} 
  \begin{tabular}{lcc|c|c|c|c}
    \hline
    \multirow{2}{*}{Method} & \multirow{2}{*}{Uncertainty} & \multirow{2}{*}{LANS} &
    \multicolumn{2}{c|}{ETH/UCY} & \multicolumn{2}{c}{SDD} \\  \cline{4-7}
    & & & ADE & FDE & ADE & FDE \\ \cline{1-7}
    \multirow{3}{*}{\textit{Diffusion$^\mathbf{2}$}} 
        & \ding{55} & \ding{55} & 0.22 & 0.37 & 8.45 & 15.59 \\ 
        & \ding{55} & \ding{51} & 0.23 & 0.38 &8.34  & 15.09  \\ 
        & \ding{51} & \ding{51} & \textbf{0.19} & \textbf{0.33} &\textbf{8.26}  &\textbf{14.87}  \\ 
    \hline
  \end{tabular}
\end{table}

We conduct ablation studies, and the results are shown in Tab.~\ref{tab:mulan_uncertainty}. Specifically, we focus on two core modules: uncertainty estimation and learnable adaptive noise scheduling (LANS).

We start by using two independent diffusion models to predict the unobserved history and future trajectories, without any additional modules. Then, we add the LANS module while keeping the uncertainty estimation disabled. On the ETH/UCY dataset, we observe a slight performance drop, suggesting that without explicit uncertainty guidance, LANS may lack meaningful modulation signals, which can impair the noise injection process and affect future trajectory quality. On the SDD dataset, however, LANS still brings mild improvement even without uncertainty, possibly due to its ability to adjust noise scale based on training dynamics in simpler open-space scenarios.

Finally, enabling both uncertainty estimation and LANS (i.e., the full \textit{Diffusion$^\mathbf{2}$}) yields the best performance on both datasets. This demonstrates that uncertainty-guided adaptive noise scheduling helps the model capture historical trajectory characteristics more effectively, leading to improved trajectory prediction accuracy.

\section{Conclusion and Discussion}

In this study, we propose a novel framework \textit{Diffusion$^\mathbf{2}$}, specifically for momentary trajectory prediction. Our proposed model consists of two sequentially connected diffusion models: one for generating unobserved historical trajectories and the other for predicting future trajectories. Considering the noise of the predicted historical trajectories, we design a dual-headed parameterization mechanism to estimate its uncertainty and a learnable adaptive noise module to dynamically adjust the noise scale in the forward diffusion process. Experiments show that our \textit{Diffusion$^\mathbf{2}$} outperforms the state-of-the-art methods on multiple datasets. Meanwhile, we also note that the diffusion-based framework has inherent limitations. In particular, its iterative sampling process leads to slow inference speed, which may hinder its deployment in real-time scenarios. In addition, since the training process involves optimizing multiple diffusion stages, it can be computationally expensive. In future work, we plan to explore more efficient training and inference methods to reduce computational resource overhead while maintaining prediction quality.

\noindent\textbf{Limitations.} 
Although \textit{Diffusion$^\mathbf{2}$} delivers strong results, it still has limitations. We observe reduced adaptability in interaction intensive scenes, such as those in the UNIV dataset. In future work we will improve efficiency and robustness, and we will validate the framework in more complex traffic scenarios.

\section*{Acknowledgement}

We thank the Pacific Northwest’s Transportation Consortium (PacTrans) Center for their funding support.

\bibliographystyle{IEEEtran}
\bibliography{refs}



\begin{IEEEbiography}[{\includegraphics[width=1in,height=1.35in,clip,keepaspectratio]{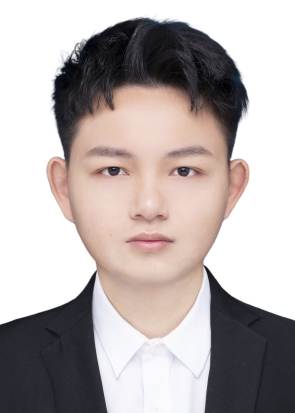}}]{Yuhao Luo} (Graduate Student Member, IEEE) received the B.S. degree in vehicle engineering from Wuhan University of Technology and the M.S. degree in Intelligent Transportation from The Hong Kong University of Science and Technology (Guangzhou), Guangzhou, China. He is currently pursuing the Ph.D. degree in Civil and Environmental Engineering with the Sky-Lab, University of Wisconsin–Madison, Madison, WI, USA. His research interests include end-to-end autonomous driving, motion and trajectory prediction, spatiotemporal machine learning, and world model.
\end{IEEEbiography} 

\begin{IEEEbiography}[{\includegraphics[width=1in,height=1.35in,clip,keepaspectratio]{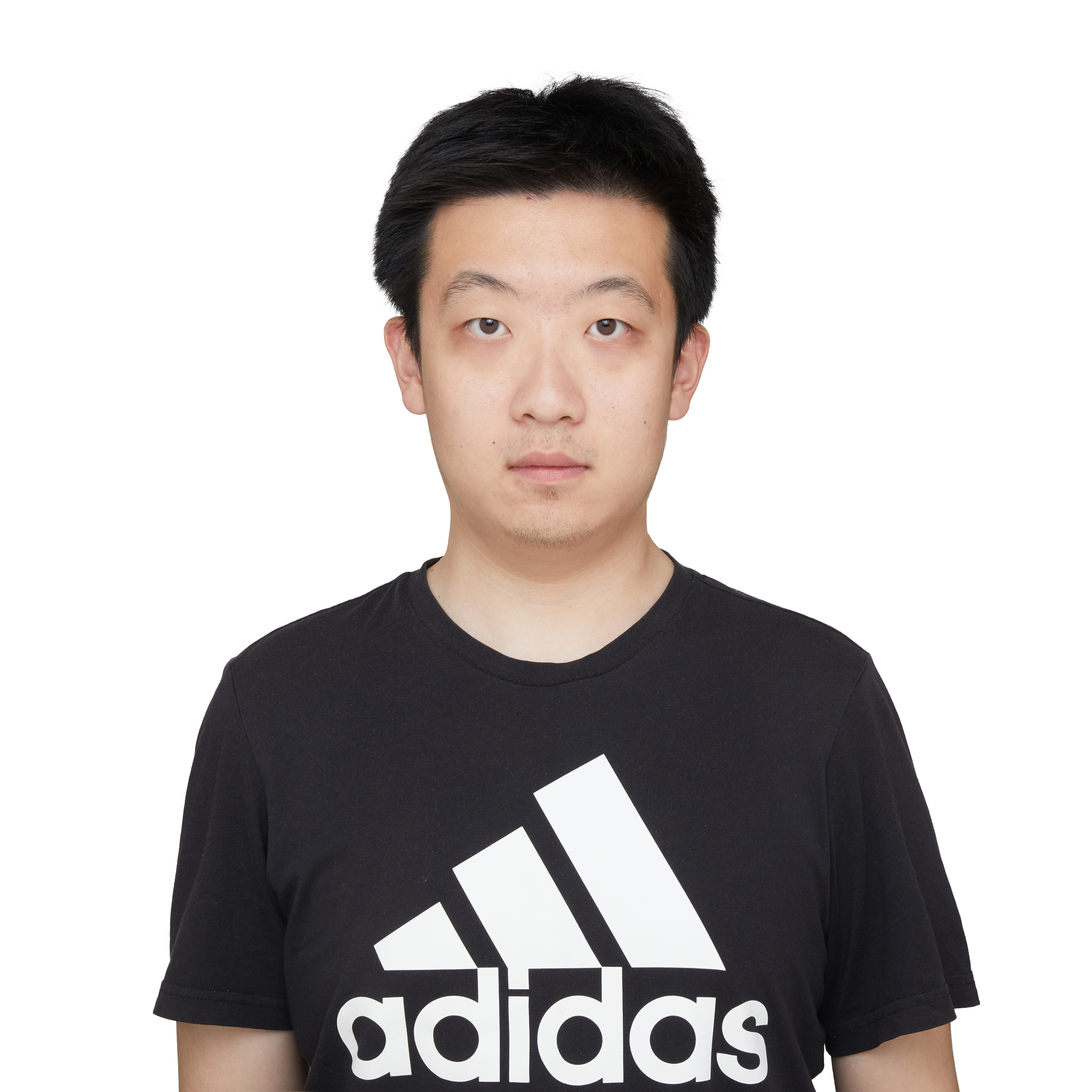}}]{Yuang Zhang} received the bachelor's and master's degrees from the Department of Automation, Tsinghua University. He is currently working toward the PhD degree with the Department of Civil and Environmental Engineering, University of Washington, Seattle,Washington. He has published papers in IEEE TRANSACTIONS ON MOBILE COMPUTING (TMC), IEEE ICRA and IEEE IST, etc. His research interests are focused on generative AI, computer vision and autonomous driving.
\end{IEEEbiography}

\begin{IEEEbiography}[{\includegraphics[width=1in,height=1.35in,clip,keepaspectratio]{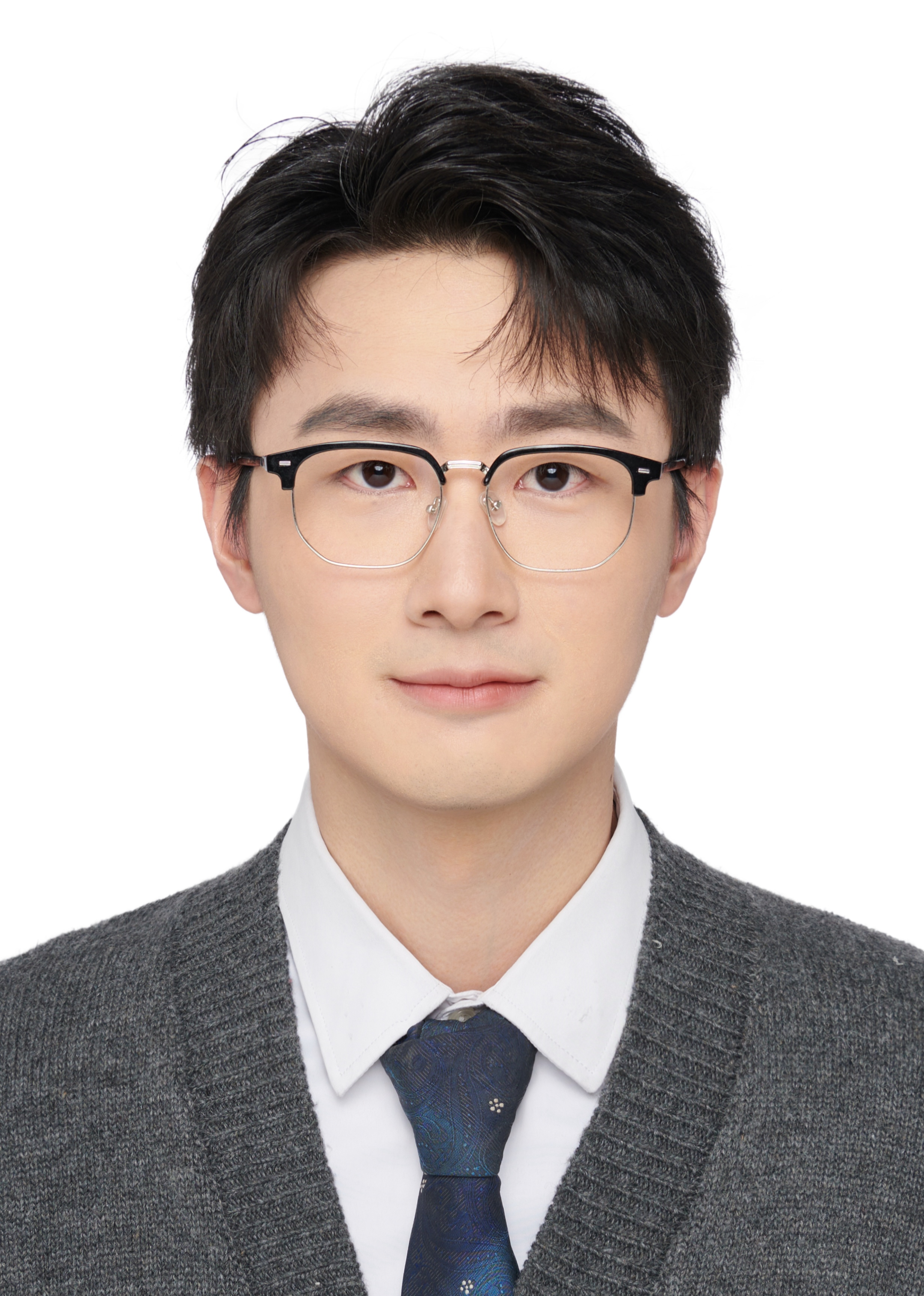}}]{Kehua Chen} (Member, IEEE) received the B.S. degree in civil engineering from Chongqing University, the dual M.S. degree in environmental sciences from the University of Chinese Academy of Sciences and the University of Copenhagen, and the Ph.D. degree in intelligent transportation from The
Hong Kong University of Science and Technology
in 2024. Currently, he is a Post-Doctoral Scholar at the Smart Transportation Applications and Research (STAR) Laboratory, University of Washington. His research interests encompass urban and sustainable
computing, and autonomous driving.
\end{IEEEbiography} 

\begin{IEEEbiography}[{\includegraphics[width=1in,height=1.35in,clip,keepaspectratio]{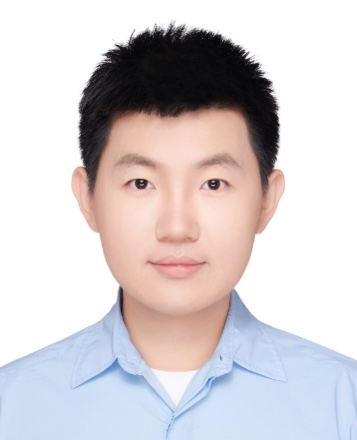}}]{Xinyu Zheng} received a MPhil degree in vehicle Intelligent Transportation from The Hong Kong university of Science and Technology (Guangzhou), Guangzhou, China, in 2025. He is currently working toward the Ph.D. degree with the College of Transportation, Tongji University, Shanghai, China. His research interests include Generating Models and Railway Safety.
\end{IEEEbiography} 

\begin{IEEEbiography}[{\includegraphics[width=1in,height=1.35in,clip,keepaspectratio]{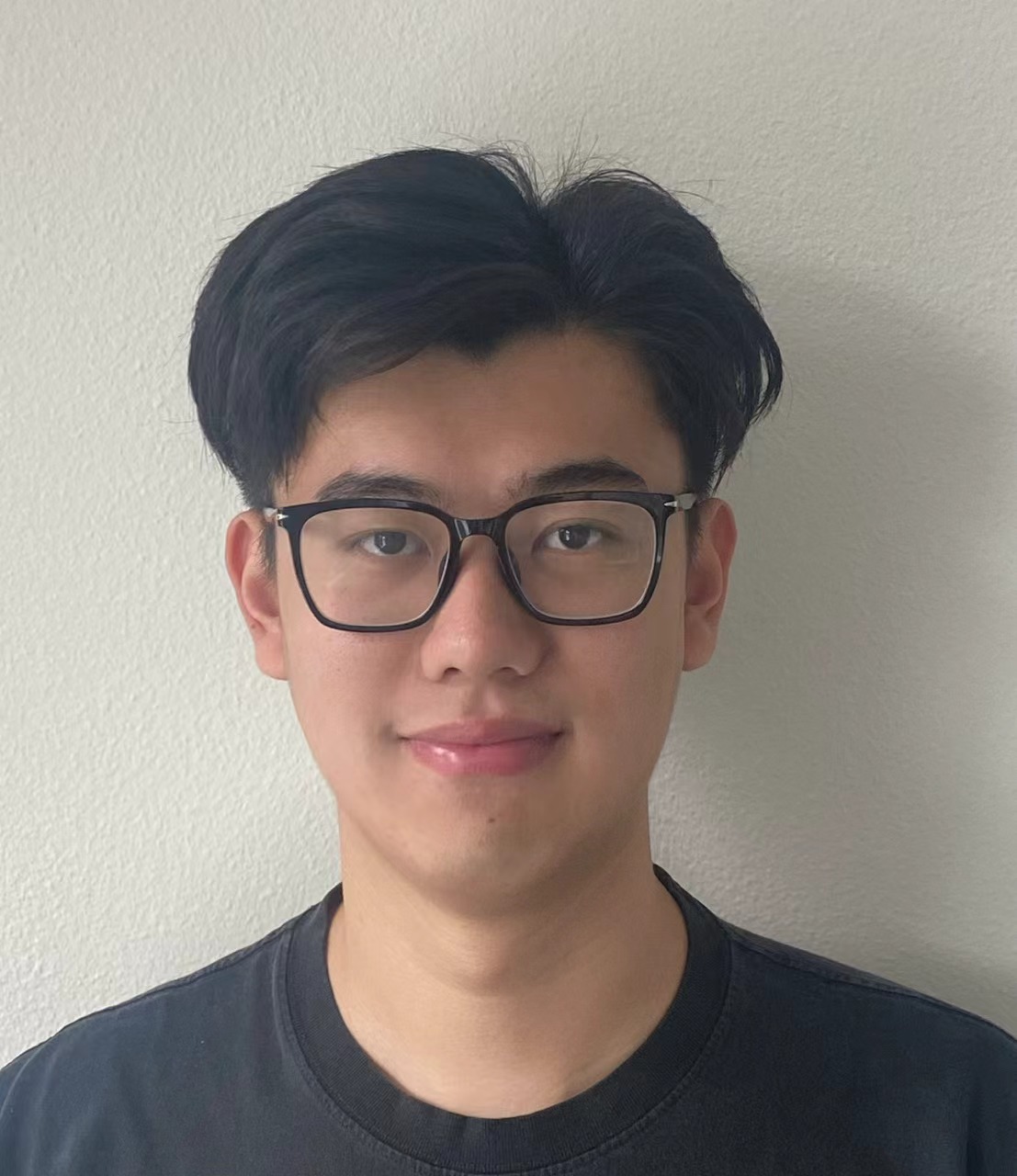}}]{Shucheng Zhang} is currently pursuing the Ph.D. degree in the Smart Transportation Application and Research (STAR) Lab, Department of Civil and Environmental Engineering at the University of Washington. Prior to joining the STAR Lab, Shucheng earned his M.S. in Mechanical Engineering from Duke University. His research interests include computer vision, intelligent transportation systems, and autonomous vehicles, with a focus on developing innovative solutions to enhance road safety and transportation automation.
\end{IEEEbiography} 

\begin{IEEEbiography}[{\includegraphics[width=1in,height=1.35in,clip,keepaspectratio]{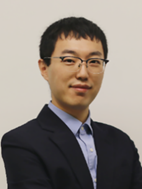}}]{Sikai Chen} is an Assistant Professor at the Department of Civil and Environmental Engineering and the Department of Mechanical Engineering (courtesy), University of Wisconsin-Madison. He received his Ph.D. in Civil Engineering with a focus on Computational Science \& Engineering from Purdue University. His research centers around three major themes: human users, AI, and transportation. He aims to innovate and develop safe, efficient, resilient, and human-centered transportation systems using cutting-edge methods and technologies. The focus is on incorporating human behaviors, interactive autonomy, digital infrastructure, and domain-specific AI. In addition, he is a member of two ASCE national committees: Connected \& Autonomous Vehicle Impacts, and Economics \& Finance; and IEEE Emerging Transportation Technology Testing Technical Committee.
\end{IEEEbiography}

\begin{IEEEbiography}[{\includegraphics[width=1in,height=1.35in,clip,keepaspectratio]{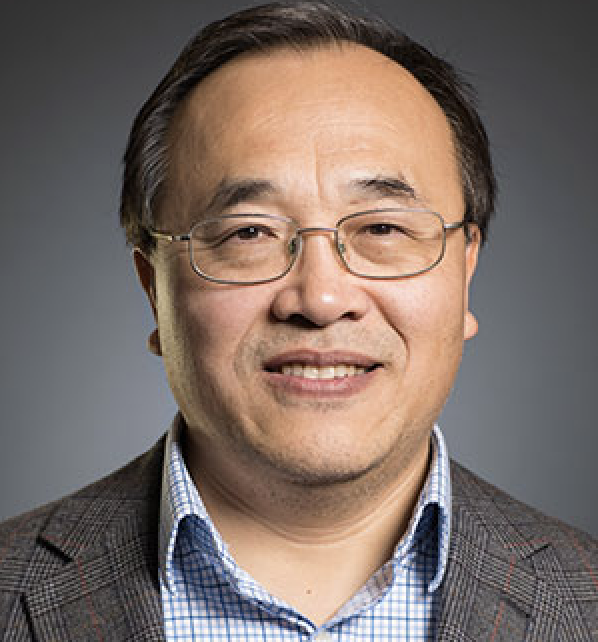}}]{Yinhai Wang} received the master’s degree in computer science from the University of Washington (UW) and the Ph.D. degree in transportation engineering from The University of Tokyo in 1998. He is currently a Professor in transportation engineering and the Founding Director of the Smart Transportation Applications and Research Laboratory (STAR Lab), UW. He also serves as the Director of the Pacific Northwest Transportation Consortium (PacTrans), U.S. Department of Transportation, University Transportation Center for Federal Region 10.
\end{IEEEbiography}

\end{document}